\PassOptionsToPackage{table}{xcolor}
\documentclass[sigconf]{acmart}
\usepackage[ruled,linesnumbered]{algorithm2e}
\usepackage{multicol,multirow}
\usepackage{algorithmic}
\usepackage{amsmath}
\usepackage[table]{xcolor}
\usepackage{booktabs}
\definecolor{mygray}{RGB}{240,240,240}
\definecolor{ACMLightBlue}{RGB}{224,240,249}
\AtBeginDocument{%
  }

\setcopyright{acmlicensed}
\copyrightyear{2018}
\acmYear{2018}
\acmDOI{XXXXXXX.XXXXXXX}

\renewcommand\footnotetextcopyrightpermission[1]{} %

\acmConference[Conference acronym 'XX]{Make sure to enter the correct
  conference title from your rights confirmation email}{June 03--05,
  2018}{Woodstock, NY}
\acmISBN{978-1-4503-XXXX-X/2018/06}

\begin{document}
\fancyhf{} %
\pagestyle{empty} %
\title{3D CoCa: Contrastive Learners are 3D Captioners}

\author{Ting Huang$^{1*}$\quad Zeyu Zhang$^{2*\dag}$\quad Yemin Wang$^{3*}$\quad Hao Tang$^{4\ddag}$\\
\vspace{0.2cm}
$^1$Shanghai University of Engineering Science\quad
$^2$The Australian National University\\
$^3$Xiamen University\quad
$^4$Peking University\\
\small $^*$Equal contribution. $^\dag$Project lead. $^\ddag$Corresponding author: bjdxtanghao@gmail.com.}

\begin{abstract}
3D captioning, which aims to describe the content of 3D scenes in natural language, remains highly challenging due to the inherent sparsity of point clouds and weak cross-modal alignment in existing methods. To address these challenges, we propose 3D CoCa, a novel unified framework that seamlessly combines contrastive vision-language learning with 3D caption generation in a single architecture. Our approach leverages a frozen CLIP vision-language backbone to provide rich semantic priors, a spatially-aware 3D scene encoder to capture geometric context, and a multi-modal decoder to generate descriptive captions. Unlike prior two-stage methods that rely on explicit object proposals, 3D CoCa jointly optimizes contrastive and captioning objectives in a shared feature space, eliminating the need for external detectors or handcrafted proposals. This joint training paradigm yields stronger spatial reasoning and richer semantic grounding by aligning 3D and textual representations. Extensive experiments on the ScanRefer and Nr3D benchmarks demonstrate that 3D CoCa significantly outperforms current state-of-the-arts by 10.2\% and 5.76\% in CIDEr@0.5IoU, respectively.
Code will be available at \url{https://github.com/AIGeeksGroup/3DCoCa}.
\end{abstract}



\maketitle

\section{Introduction}
In recent years, 3D learning research has been increasing, driven by various practical applications such as robotics, autonomous driving, and augmented reality~\cite{sportscap2021,chen2021tightcap,Liao_Zhu_Zhang_Ye_Chen_Fan_2021,dcnet2023}. Within this burgeoning field, the intersection of computer vision (CV) and natural language processing (NLP) has prompted researchers to strive to bridge the gap between visual perception and language expression, thus promoting the rise of cross-modal tasks such as visual captioning. The emergence of large-scale vision-language models has brought unprecedented breakthroughs in the generation of captions for 2D images. With the development of 3D vision-language datasets, 3D captions have also shown promising prospects. 3D captioning extends 2D image captioning and aims to accurately perceive the 3D structure of objects and generate reasonable descriptions by leveraging a comprehensive set of attribute details and contextual interaction information between objects and their surroundings. However, due to the sparsity of point clouds and the cluttered distribution of objects, describing objects within a 3D scene remains a particularly challenging endeavor.
\begin{figure}[t]
    \centering
    \includegraphics[width=\linewidth]{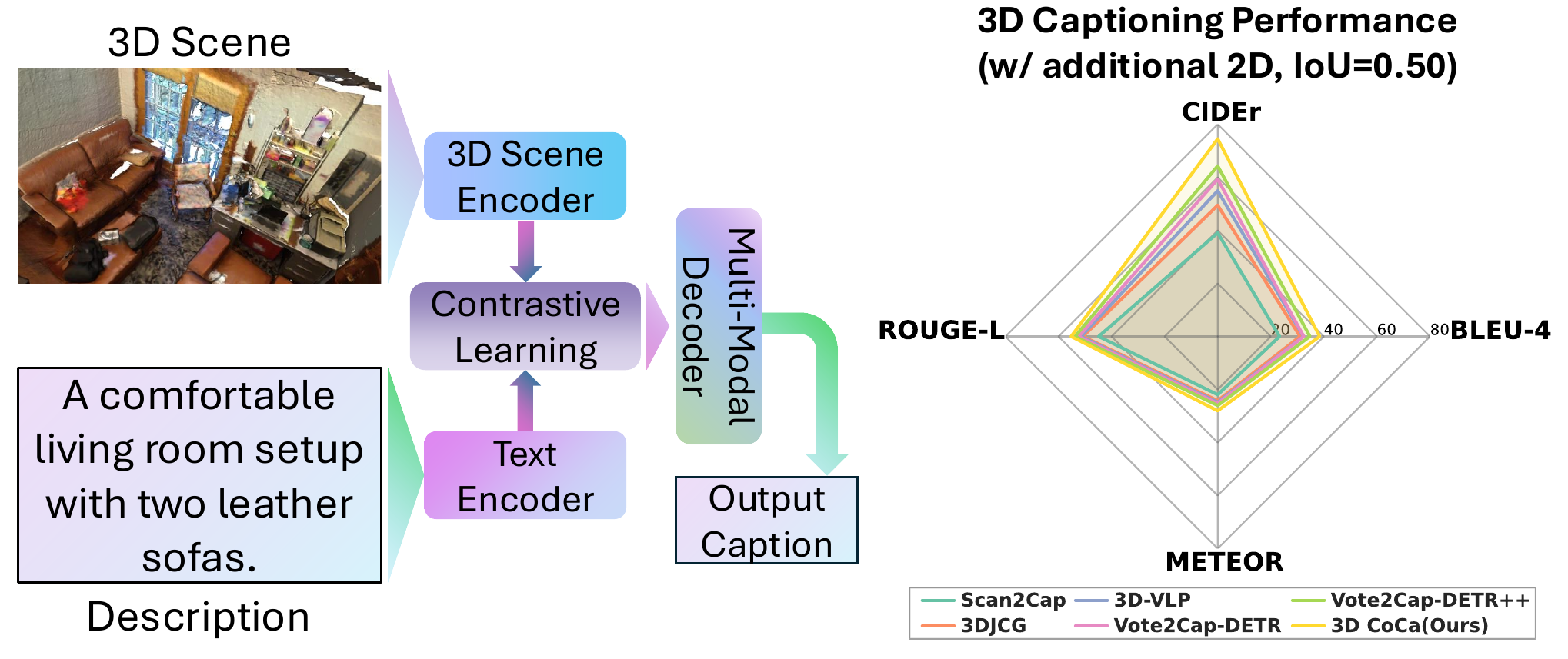}
    \caption{Conceptual homepage figure for 3D CoCa, highlighting its architecture (left) and performance (right). Left: The 3D CoCa model unifies contrastive learning and multimodal captioning in one framework. Right:Radar chart comparison of 3D CoCa and previous methods Scan2Cap~\cite{scan2cap_2021}, 3DJCG~\cite{3djcg2022}, 3D-VLP~\cite{3dvlp2024}, Vote2Cap-DETR~\cite{vote2cap2023}, Vote2Cap-DETR++~\cite{vote2cap++2024} on the ScanRefer~\cite{chen2020scanrefer} benchmark.}
	\label{fig:main}
    \vspace{-0.4cm}
\end{figure}

Early approaches to 3D dense captioning adopted a two-stage ``detect-then-describe'' paradigm, where object proposals were first detected from point clouds and then described individually. For example, Scan2Cap~\cite{scan2cap_2021} is the first attempt to integrate 3D object detection and caption generation into 3D scenes in a cascade manner. \cite{camm2023} introduced a novel 3D language pre-training approach that uses context-aware alignment and mutual masking to learn generic representations for 3D dense captioning tasks. Although effective, a two-stage pipeline can suffer from significant performance degradation. First, the detection stage usually produces redundant bounding boxes, and thus careful tuning using the Non-Maximum Suppression (NMS)~\cite{nms2006} operation is required, which introduces additional hyperparameters and increases computational overhead. Second, the cascade design of the ``detect-then-describe'' process makes caption generation highly dependent on the quality of the detection stage. In this context, the exploration of one-stage end-to-end 3D dense captioning models has attracted widespread attention. Vote2Cap-DETR~\cite{vote2cap2023} and its advanced version Vote2Cap-DETR++~\cite{vote2cap++2024} are notable examples, using the Transformer framework to simultaneously locate and describe objects during inference in a single forward pass, improving both efficiency and performance. Other recent approaches, such as BiCA~\cite{bica2025} introduced a Bi-directional Contextual Attention mechanism to disentangle object localization from contextual feature aggregation in 3D scenes and See-It-All (SIA) model~\cite{seeitall2024} adopted a late aggregation strategy to capture both local object details and global contextual information with a novel aggregator. Moreover, TOD3Cap~\cite{todocap2025} employed a Bird’s Eye View (BEV) representation for the generation of object proposals and integrated the Q-Former Relation with the LLaMA-Adapter to generate descriptive sentences, particularly for outdoor environments.

Despite progress, 3D captioning remains very challenging, especially in modeling spatial relations and aligning 3D visual data with textual semantics. Describing complex spatial arrangements requires the model to understand 3D geometry and relative object positions, which is non-trivial to encode and reason about. Bridging the gap between the 3D modality and language is also difficult. Existing methods treat vision and language as separate stages with weak cross-modal interaction. This leads to suboptimal alignment between visual and textual representations.

These challenges point to the need for a unified framework that can enhance spatial reasoning and cross-modal alignment using strong visual-linguistic priors. Foundation models in vision-language research CoCa~\cite{coca2022} have shown that contrastive pre-training on large image-text corpora yields representations with rich semantics and excellent alignment between modalities. Inspired by this, we hypothesize that bringing such powerful priors into 3D captioning will significantly improve performance and generalization. This insight motivates us to design a 3D captioning approach that jointly learns spatially-grounded captions and visual-text alignments within a single end-to-end model, leveraging knowledge from large-scale vision-language training.

In this paper, we introduce 3D CoCa (Contrastive Captioner for 3D), as illustrated in Figure~\ref{fig:main}, a novel approach that integrates contrastive learning and caption generation into a unified model for 3D scenes. The core idea is to train a 3D scene encoder and a text encoder together with a shared contrastive learning objective, while simultaneously training a multi-modal decoder to generate captions. By coupling these tasks, 3D CoCa learns a joint feature space where 3D representations and captions are deeply aligned. The model leverages rich semantic knowledge from large-scale pre-training: we build on a vision-language backbone initialized with learned visual and linguistic features, injecting strong priors about objects and language into the 3D domain. This allows the model to recognize a wide range of concepts in the scene and associate them with the correct words. Furthermore, 3D CoCa is designed to be spatially aware – the 3D scene encoder preserves geometric structure, and the decoder’s attention mechanism can attend to specific regions when wording the description. As a result, the generated captions capture not only object attributes, but also their precise spatial context, directly addressing the core difficulty of 3D captioning. In essence, our approach marries a powerful contrastive learner with a captioning model, demonstrating that contrastive learners are effective 3D captioners.

In summary, the main contributions of this work include:
\begin{itemize} 
\setlength\itemsep{0em}
    \item We propose 3D CoCa, the first end-to-end framework to unify contrastive vision-language learning with 3D captioning. This design eliminates the need for external 3D object detectors by jointly learning to localize and describe from point clouds.
    
    \item We demonstrate how to leverage strong visual-linguistic priors from large-scale image-text pretraining within a 3D captioner. By integrating a contrastive alignment objective, our model attains improved semantic understanding and cross-modal alignment, enabling richer and more accurate captions for complex 3D scenes.

    \item Extensive evaluations on benchmark datasets show that 3D CoCa achieves state-of-the-art captioning performance on Nr3D~\cite{achlioptas2020referit_3d} (52.84\% C@0.5) and Scanrefer~\cite{chen2020scanrefer} (77.13\% C@0.5).
	
\end{itemize} 

\begin{figure*}[t]
    \centering
    \includegraphics[width=\linewidth]{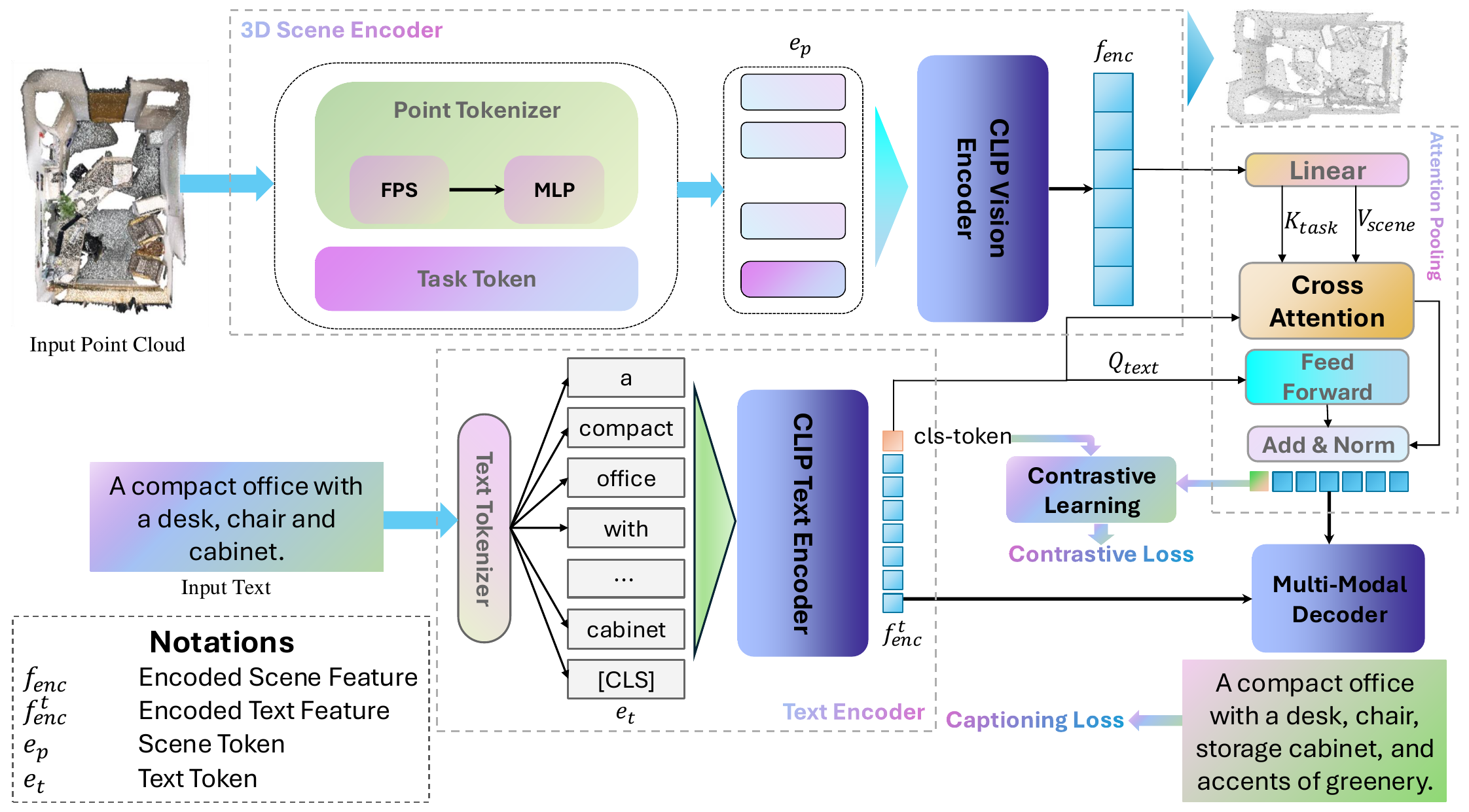}
    \caption{Illustration of a multi-modal Transformer architecture for 3D vision-language understanding. The input point cloud and textual description are processed by CLIP Vision and Text Encoders, respectively. Cross-attention mechanisms fuse these features within a Multi-Modal Decoder, enabling the generation of descriptive captions. The model training is guided by contrastive and captioning losses, promoting effective alignment between visual and textual modalities.}
	\label{fig:pipeline}
    \vspace{-0.4cm}
\end{figure*}

\section{Related Works}
\paragraph{3D Dense Captioning}
3D dense captioning involves localizing objects in a 3D scene and describing them in natural language. Early work like Scan2Cap~\cite{scan2cap_2021} pioneered this task by leveraging point cloud data with spatial reasoning, marking a departure from conventional 3D detection pipelines focused only on classification and bounding boxes \cite{zhao2025peddet,zhang2024meddet,cai2024medical,cai2024msdet}. Subsequent methods were built on this foundation with improved relational modeling. For example, the Multi-Order Relation Extraction (MORE) framework~\cite{MORE_2022} introduced higher-order relationship reasoning, showing that richer spatial context leads to more informative and accurate captions.

The introduction of Transformer architectures further accelerated progress in 3D captioning. SpaCap3D~\cite{spa2cap2022} employed a Transformer-based encoder–decoder with a spatially guided encoder to capture geometric context and an object-centric decoder for attribute-rich descriptions. $\chi$-Trans2Cap~\cite{trans2cap2022} extended this idea by distilling knowledge from 2D vision-language models into a 3D captioner, effectively transferring semantic understanding from images to point clouds. Recent works strive for unified architectures that handle multiple tasks: 3DJCG~\cite{3djcg2022} uses shared Transformers to jointly optimize 3D captioning and visual grounding, and UniT3D~\cite{unit3d2023} demonstrates that pre-training a Transformer on large-scale point cloud–text pairs can yield state-of-the-art results across diverse 3D scene understanding benchmarks.

Despite these advances, most approaches still follow a two-stage ``detect-then-describe'' paradigm~\cite{scan2cap_2021,trans2cap2022,3djcg2022,spa2cap2022}, where an object detector provides regions that are then described. This separation can cause error propagation and misalignment between the vision and language components. To overcome this limitation, end-to-end paradigms have been explored. Vote2Cap-DETR~\cite{vote2cap2023} and its improved variant Vote2Cap-DETR++~\cite{vote2cap++2024} reformulate dense captioning as a direct set-prediction task, similar to DETR in 2D vision. They jointly localize and caption objects in one stage, eliminating dependence on pre-trained detectors. Through a Transformer encoder–decoder with learnable queries and iterative refinement, these one-stage models achieve competitive performance while simplifying the pipeline.

\paragraph{3D Pre-training and Vision-Language Foundations}
Another line of work has focused on pre-training 3D representations to provide stronger foundations for downstream tasks. Unsupervised 3D representation learning techniques can be categorized into global contrastive methods~\cite{Wang_Liu_Yue_Lasenby_Kusner_2021,Mei_Huang_Liu_Zhang_Wu_2022} that learn holistic point cloud embeddings, local contrastive methods~\cite{PointContrast2020,wang2023tap} that distinguish fine-grained geometric structures or multi-view correspondences, and masked point modeling approaches~\cite{yu2021pointbert,pang2022masked} that adapt masked autoencoding to 3D data. These approaches learn powerful geometric features; however, they operate purely on 3D geometry and lack grounding in natural language semantics.

To bridge this gap, researchers have explored 3D vision-language pre-training. For example, 3D-VLP~\cite{3dvlp2024} uses contrastive learning to align point cloud segments with text descriptions, yielding representations that improve 3D dense captioning and visual grounding performance by injecting semantic knowledge. Similarly, UniT3D~\cite{unit3d2023} showed that training on large-scale point cloud–caption pairs endows a unified model with strong multi-task 3D understanding capabilities. Such findings underscore the value of learning joint 3D–language representations as a foundation for captioning.

\paragraph{Multimodal Large Language Models for 3D Scenes}
Recently, the success of large language models in vision-language tasks has sparked interest in extending them to 3D scene understanding. A representative example is LL3DA~\cite{ll3da2024}, a ``Large Language 3D Assistant'' that combines 3D visual inputs with an LLM, allowing the model to follow natural-language instructions and generate responses about a 3D scene. This enables interactive tasks such as 3D captioning, visual grounding, and question answering by leveraging the reasoning ability of LLMs. Similarly, Chat-3D~\cite{huang2024chat} aligns point cloud features directly with a pretrained language model’s embedding space, demonstrating impressive conversational capabilities to describe 3D environments. Such systems illustrate the promise of MLLMs for 3D grounding, dense captioning, and dialogue-based interaction.

However, these LLM-driven frameworks typically rely on an external language model and complex alignment procedures, treating captioning as just one of many tasks rather than a dedicated end-to-end objective. Consequently, fine-grained spatial details can be difficult to handle without additional tricks. In contrast, 3D CoCa takes a different route: it directly integrates multimodal pre-training into a unified captioning architecture. By jointly training a 3D scene encoder and a text decoder with a contrastive vision-language objective, 3D CoCa harnesses rich semantic priors from foundation models while remaining end-to-end trainable for the captioning task. This design eliminates the need for separate detection modules or post-hoc LLM integration; to our knowledge, 3D CoCa is the first to unify contrastive vision-language pre-training with 3D dense captioning in a single model, marking a novel paradigm in 3D captioning within the evolving MLLM-centered landscape.

\vspace{-0.4cm}

\section{The Proposed Method}
\label{sec:method}
\subsection{Overview}
In this section, we present the proposed 3D CoCa, a framework that bridges the gap between 3D point cloud representation learning and natural language understanding for captioning. Our approach builds on principles of contrastive alignment and multi-modal captioning, inspired by the successes of CLIP-style image-text models~\cite{clip2021} and the Contrastive Captioner (CoCa) paradigm~\cite{coca2022}. As illustrated in Figure~\ref{fig:pipeline}, 3D CoCa consists of four key components: a 3D Scene Encoder, a Text Encoder, a Contrastive Learning module, and a Multi-Modal Fusion Decoder.

Unlike traditional methods that focus on either 2D images or purely 3D data, 3D CoCa leverages knowledge distilled from large-scale 2D-text pre-training and adapts it to the complexities of point cloud data. Most of CLIP’s pre-trained weights are frozen in our framework to preserve the robust visual and linguistic representations, introducing only minimal additional parameters for 3D processing. The following subsections describe each component in detail. We conclude this section with the joint training objectives that bind these components into a unified model for generating captions from 3D scenes.

\vspace{-0.4cm}

\subsection{3D Scene Encoder}
\label{subsec:sceneencoder}
The role of the 3D scene encoder is to transform an unstructured point cloud into a set of latent tokens that capture the scene’s geometric and semantic content. We build our scene encoder based on the EPCL architecture~\cite{huangepcl}, a design that integrates point-based processing with a frozen 2D CLIP visual backbone. The encoder comprises three parts: (i) a point cloud tokenizer that groups points into patch tokens, (ii) a set of learnable task tokens that inject 3D-captioning context, and (iii) a frozen CLIP vision transformer that encodes the combined token sequence. Figure~\ref{fig:pipeline} (top-left) depicts how raw point clouds are converted into tokens and fed into the encoder.
\subsubsection{Point cloud tokenizer}
Given an input point cloud $P \in \mathbb{R}^{N \times (3+F)}$ (with $N$ points, each described by 3D coordinates $(x,y,z)$ and $F$ additional features such as color, normal, height or multiview feature), we first convert it into a discrete token sequence. We sample $M$ representative points as patch centers using farthest point sampling (FPS) to ensure even coverage of the scene. FPS reduces redundancy in dense regions while preserving structure in sparse areas. Next, for each sampled center, we group its $K$ nearest neighbor points to form a local patch. This yields $M$ patches ${P_1, P_2, \dots, P_M}$, each containing $K$ points that are spatially proximate. We then pass each patch through a small point-wise network (a series of Multi-Layer Perceptrons, MLPs) to encode local geometry and appearance features. This produces a set of $M$ point tokens (one per patch), each a $D_p$-dimensional embedding:
\begin{equation}
E_p(P) = [\mathbf{e}_{p_1}, \mathbf{e}_{p_2}, \dots, \mathbf{e}_{p_M}]\in\mathbb{R}^{M \times D_p},
\end{equation}
where $\mathbf{e}_{p_i}$ is the embedding of the $i$-th patch. By treating each local patch as a token, the continuous 3D data is converted into a structured sequence of vectors. This tokenization balances fine local detail (within each patch of $K$ points) and global coverage (through the $M$ sampled patches) of the scene.

\subsubsection{Task token mechanism}
While the above point tokens capture visual elements of the scene, the model still needs guidance that the task is 3D captioning (describing the scene in words). To provide this context, we introduce a small set of learnable task tokens. Each task token is an embedding vector (implemented as part of the model parameters) that is prepended to the sequence of point tokens. Following the prompt tuning approach in~\cite{ptuning2022}, we initialize these task token embeddings with distinct fixed values (e.g. enumerated numbers) and allow them to be learned. The task tokens act as a high-level prompt or query that informs the model about the captioning task. By attending over the entire point cloud, these tokens learn to pull out global semantic information (e.g. the overall scene context or salient objects) that is useful for generating descriptive text. In essence, the task tokens provide a shared contextual bias for the 3D scene, helping the encoder emphasize elements relevant to language description.

\subsubsection{Frozen CLIP vision encoder}
After obtaining the $M$ point tokens and $m_t$ task tokens, we concatenate them into a single sequence:
\begin{equation}
    [\mathbf{e}_{p_1}, \dots, \mathbf{e}_{p_M}; \mathbf{t}_1, \dots, \mathbf{t}_{m_t}],
\end{equation}
where $\mathbf{t}_j$ denotes the $j$-th task token embedding. This combined sequence of length $M + m_t$ is then fed into the CLIP visual Transformer encoder~\cite{clip2021}. We adopt the CLIP image encoder architecture and keep its weights frozen to leverage the rich visual features it learned from massive image-text data. Freezing the CLIP vision backbone preserves its robust representation power and stabilizes training – we avoid updating a large number of parameters, thus preventing ``catastrophic forgetting'' of prior knowledge. It also improves efficiency: with most parameters fixed, memory usage, and training time are significantly reduced.

The CLIP vision encoder processes the token sequence and outputs a sequence of latent features in a high-dimensional space. This output encodes both the 3D geometry and the task context. From these outputs, we can derive a global scene representation that will be used for downstream alignment with text. In practice, we obtain the global 3D scene feature $f_{enc}$ from the CLIP encoder’s output. This feature $f_{enc} \in \mathbb{R}^{D}$ with $D$ the encoder output dimension is a compact, semantically rich summary of the entire 3D scene conditioned on the captioning task. It encapsulates the visual content in a form suitable for aligning with language and will serve as the 3D scene embedding for the contrastive learning module.

\subsection{Text Encoder}
While the 3D scene encoder encodes visual information from point clouds, the text encoder processes natural language descriptions into a compatible embedding space. We use the text encoder branch of CLIP~\cite{clip2021} to obtain language features. This text encoder is a Transformer-based model that we also keep frozen, so as to exploit the linguistic knowledge gained from large-scale pre-training. By using a fixed pre-trained text encoder, we ensure that our captions are encoded in the same semantic space as the CLIP representations, which facilitates alignment with the 3D scene features.
\subsubsection{Text tokenizer}
Given an input sentence $T$, we first tokenize it into a sequence of $L$ tokens. Each token $w_i$ is mapped to an embedding vector in $\mathbb{R}^{D_t}$ using a learned embedding table. This produces a sequence of text token embeddings:
\begin{equation}
    E_t(T) = [\mathbf{e}_{t_1}, \mathbf{e}_{t_2}, \dots, \mathbf{e}_{t_L},] \in \mathbb{R}^{L \times D_t},
\end{equation}
where $\mathbf{e}_{t_i}$ corresponds to the $i$-th token in the sentence. We prepend a special beginning-of-sequence token to this sequence, which will be used to aggregate the sentence-level information. We also add positional encodings to each token embedding $E_t(T)$ to preserve the order of words, which is crucial to capture the syntactic structure and meaning of the caption. We employ a subword tokenizer to handle out-of-vocabulary words by breaking them into known subunits, ensuring that any arbitrary caption can be represented by the token sequence.
\subsubsection{Frozen CLIP text encoder}
The sequence of text embeddings $E_t(T)$ is then passed through the CLIP text Transformer encoder, which has $N_{te}$ layers of multi-head self-attention and feed-forward networks. We denote the hidden states at layer $l$ as $H^l$ with $H^0 = E_t(T)$ being the input. The Transformer applies its layers successively:
\begin{equation}
    H^l = \mathrm{TransformerBlock}^l(H^{l-1}), l \in [1,\dots,N{te}],
\end{equation}
comprising self-attention, layer normalization, and MLP sublayers in each block. We keep all weights of this text encoder frozen during training to preserve the rich language understanding it acquired through pre-training on image-text pairs. Freezing also mitigates overfitting, given that 3D captioning datasets are relatively small compared to general text corpora.

From the final layer of the text Transformer, we extract the output corresponding to the special [CLS] token, which we treat as the global text representation for the caption. Denote this vector as $f_{enc}^t \in \mathbb{R}^{D_t}$. This vector encodes the semantic content of the entire description $T$ in a single feature. It will be used in our contrastive learning module to align with the 3D scene feature $f_{enc}$ from the scene encoder. By using CLIP’s text encoder and keeping it fixed, we ensure $f_{enc}^t$ lies in a language embedding space that is directly comparable to CLIP visual features, aiding the multimodal alignment.

\subsection{Contrastive Learning Paradigm}
\label{subsec:conlp}
To bridge the heterogeneous modalities of 3D point clouds and text, we adopt a contrastive learning strategy for feature alignment. The core idea is to project both the 3D scene feature $f_{enc}$ and the text feature $f_{enc}^t$ into a shared latent space where corresponding 3D-scenes and captions are pulled closer together, while non-matching pairs are pushed farther apart. This follows in the same spirit as the CLIP multimodal training objective, encouraging the model to learn cross-modal associations. We describe the feature projection and normalization, followed by the contrastive loss formulation.

\subsubsection{Feature alignment}
Before computing the similarity between $f_{enc}$ and $f_{enc}^t$, we transform them into a common embedding space using learnable projection heads. In particular, we apply two small MLPs to map the features to a shared dimension. Specifically, we use a two-layer MLP to project the features:
\begin{equation}
    \tilde{f}_{enc}=\mathrm{MLP}_v\left ( f_{enc} \right ),\qquad\tilde{f}_{enc}^t=\mathrm{MLP}_t\left ( f_{enc}^t \right ),
\end{equation}
where $\mathrm{MLP}_v$ and $\mathrm{MLP}_t$ are two-layer perceptrons for the 3D scene feature and text feature respectively. Each MLP consists of a linear layer, a ReLU activation, and a second linear layer. These learned projections ensure that the 3D and text embeddings are not only of the same dimension but also tuned for maximal alignment. After projection, we L2-normalize each feature vector to unit length:
\begin{equation}
    \hat{f}_{enc}=\frac{\tilde{f}_{enc} }{\left \| \tilde{f}_{enc}  \right \|_2 } ,\qquad \hat{f} _{enc}^t=\frac{\tilde{f}_{enc}^t }{\left \| \tilde{f}_{enc}^t  \right \|_2 } .
\end{equation}
This normalization enables direct comparison via cosine similarity during loss computation.

\subsubsection{Contrastive loss function}
With the features projected and normalized, we employ a contrastive loss to train the model to align the correct 3D-text pairs. We follow the InfoNCE loss formulation popularized by CLIP. Consider a training batch of $N$ pairs of 3D scenes and their corresponding captions. We first compute the pairwise cosine similarities between all scene–caption pairs in the batch. For the $i$-th scene and $j$-th text in the batch, the similarity is defined as:
\begin{equation}
    \mathrm{sim}\left ( \hat{f}_{enc,i},\hat{f}_{enc,j}^t   \right )  =\frac{\hat{f}_{enc,i}\cdot \hat{f}_{enc,j}^t  }{\left \| \hat{f}_{enc,i}  \right \| \left \| \hat{f}_{enc,j}^t  \right \|}, 
\end{equation}
which is simply the dot product of the two unit-normalized feature vectors. The contrastive learning objective then maximizes the similarity of each scene with its matched caption (where $i=j$) while minimizing its similarity with unmatched captions ($i \neq j$). Specifically, for each scene $i$, we define the contrastive loss using a softmax over the $N$ captions:
\begin{equation}\label{losscon}
    \mathcal{L}_{\mathrm{Con}} =-\frac{1}{N} \sum_{i=1}^{N}\log \frac{\exp \left ( \mathrm{sim}\bigl ( \hat{f}_{enc,i},\hat{f}_{enc,i}^t   \bigr )/\tau   \right )  }{ {\textstyle \sum_{j=1}^{N}}\exp \left (  \mathrm{sim}\bigl ( \hat{f}_{enc,i},\hat{f}_{enc,j}^t   \bigr )/\tau\right )  } ,
\end{equation}
where $\tau$ is a learnable temperature parameter that scales the logits before softmax. This InfoNCE loss encourages $\mathrm{sim}(f_{enc,i}, f_{enc,i}^t)$ to be larger than $\mathrm{sim}(f_{enc,i}, f_{enc,j}^t)$ for any $j\neq i$, thereby aligning the $i$-th 3D scene only with its correct description. In summary, the contrastive loss $\mathcal{L}_{\mathrm{Con}}$ provides a strong supervisory signal that couples the 3D scene features and text features, driving the model to produce a joint embedding space where cross-modal correspondences are captured.

\subsection{Multi-Modal Fusion Decoder}
The final component of 3D CoCa is the multi-modal fusion decoder, which generates natural language descriptions for the input 3D scene. This decoder takes the aligned 3D-text representations and fuses them to produce fluent, contextually grounded sentences. We design the decoder as an autoregressive Transformer that uses cross-attention to incorporate visual context at each step of generation. In essence, the decoder serves as a conditional language model: it outputs a caption word-by-word, while attending to the 3D scene features to ensure the caption accurately describes the scene. By leveraging the aligned features from the contrastive stage, the decoder can inject detailed 3D scene information into the generation process, producing descriptions that are both coherent and faithful to the visual input.

The decoder operates in an autoregressive manner. It begins with a special start-of-sequence token and generates the caption one token at a time. At each time step $t$, the decoder has access to all previously generated words $y_{<t}$ as context, and predicts the next word $y_t$. This causal self-attention mechanism within the decoder allows it to capture intra-sentence dependencies, ensuring that the resulting sentence is grammatically correct and contextually consistent. In parallel, at every decoding step, the decoder is conditioned on the 3D scene representation, so that what it writes is grounded in the scene content. We achieve this through a cross-attention mechanism.

\subsubsection{Cross-Attention mechanism}
To integrate visual information from the 3D scene into the captioning process, the decoder incorporates cross-modal attention layers. In each decoder layer, a cross-attention layer allows the decoder to attend to the encoded 3D scene tokens (the output of the 3D scene encoder from Section~\ref{subsec:sceneencoder}). Formally, let $Q_{\mathrm{text}}$ be the query matrix containing the decoder’s current hidden states (for each position in the sequence at that layer), and let $K_{\mathrm{task}}$ and $V_{\mathrm{scene}}$ be the key and value matrices derived from the set of 3D scene token embeddings. The cross-attention is computed as:
\begin{equation}
    \mathrm{Attention}\left (  Q_{\mathrm{text}},K_{\mathrm{task}},V_{\mathrm{scene}}\right )  =\text{softmax} \left ( \frac{Q_{\mathrm{text}}K_{\mathrm{task}}^{\top}}{\sqrt[]{d_k} }  \right )V_{\mathrm{scene}} ,
\end{equation}
where $d_k$ is the dimensionality of the keys. This operation produces an attention output for the decoder at each position, which is essentially a weighted sum of the 3D scene value vectors $V_{\mathrm{scene}}$, with weights determined by the compatibility of queries $Q_{\mathrm{text}}$ with keys $K_{\mathrm{task}}$. In this way, the decoder can retrieve the relevant visual information needed to accurately describe that object. The cross-attention mechanism ensures that the caption not only reflects the overall context of the scene, but also captures important local details by looking at the appropriate regions in the 3D data.

The cross-attention layers are interleaved with the self-attention layers in the decoder, allowing for a continuous exchange of information between the textual and visual modalities. This iterative process of fusing self-attention and cross-attention enables the model to build a refined understanding of the scene context while preserving the grammatical and sequential coherence of the generated text.

\subsubsection{Training objectives and joint optimization}
Training the multi-modal decoder is accomplished with a combination of captioning loss and the previously introduced contrastive loss. We jointly optimize these objectives so that the model learns to generate accurate captions and maintain cross-modal alignment at the same time. The contrastive loss $\mathcal{L}_{\mathrm{Con}}$ (Eq.~\eqref{losscon}) applied to the encoder outputs encourages the 3D and text features to stay aligned, which provides a good initialization and constraint for the decoder’s cross-attention. Meanwhile, the decoder itself is primarily supervised by a captioning loss that measures how well its generated text matches the reference description.

For the caption generation task, we use the standard cross-entropy loss between the predicted caption and the ground-truth caption. Given a generated caption $\hat{Y}=\left ( \hat{y}_1,\hat{y}_2,\cdots ,\hat{y}_L    \right )  $ and the corresponding ground truth $Y=\left ( y_1,\cdots ,y_L    \right )  $, the captioning loss is defined as:
\begin{equation}
    \mathcal{L}_{\mathrm{Cap}}=-\sum_{t=1}^{L} \log P\left ( \hat{y}_t=y_t\mid \hat{y}_{< t} ,f_{enc}  \right ) ,
\end{equation}
where $f_{enc}$ is the conditioning global 3D feature, ensuring that the generated sentence is tightly linked to the visual content of the scene.

This captioning loss is jointly optimized with the contrastive loss described in the previous section \ref{subsec:conlp}. The total loss function is expressed as:
\begin{equation}
    \mathcal{L}_{\mathrm{Total}}=\mathcal{L}_{\mathrm{Con}}+\lambda \cdot \mathcal{L}_{\mathrm{Cap}},
\end{equation}
where $\lambda$ is a scalar hyperparameter that balances the two terms. By tuning $\lambda$, we regulate the trade-off between enforcing multimodal feature alignment and producing accurate natural-language output. In our experiments, we set $\lambda$ to give roughly the same importance to both objectives.

This joint optimization scheme causes the two parts of the model to reinforce each other. The contrastive alignment ensures that the visual encoder produces features that are readily attended to by the text decoder. Conversely, the act of captioning provides feedback that can refine the shared embedding space - the decoder will only succeed if the visual features $f_{enc}$ encode the information needed for generation, which in turn pressures the encoder to capture fine-grained, caption-relevant details. Overall, the combined loss drives the model to generate captions that are not only linguistically fluent and descriptive, but also correspond closely to the 3D scene content. Jointly training 3D CoCa in this manner leads to improved integration of visual context into the language output, and a tighter cross-modal correspondence between the 3D scenes and their generated captions.

\vspace{-0.3cm}

\begin{algorithm}[h]
\caption{3D CoCa Algorithm}
\label{alg:3dcoca}
\begin{algorithmic}[1]
\REQUIRE Point cloud data $P$, Text input $T$
\ENSURE Generated caption $\hat{C}$

\STATE \textbf{Point Cloud \& Text Input Processing:}
    \STATE $\mathbf{E}_p \leftarrow \text{Point cloud tokenizer}(P)$ 
        \COMMENT{Tokenize input point cloud into sequence}
    \STATE $\mathbf{E}_t \leftarrow \text{Text tokenizer}(T)$ 
        \COMMENT{Tokenize input text into sequence}

\STATE \textbf{Feature Encoding via Frozen CLIP Encoders:}
    \STATE $\mathbf{f}_{enc} \leftarrow \text{CLIP}_{\text{visual}}(\mathbf{E}_p)$ 
        \COMMENT{Frozen CLIP visual encoder}
    \STATE $\mathbf{f}_{enc}^t \leftarrow \text{CLIP}_{\text{text}}(\mathbf{E}_t)$ 
        \COMMENT{Frozen CLIP text encoder}

\STATE \textbf{Feature Alignment \& Contrastive Learning:}
    \STATE $(\hat{\mathbf{f}}_{enc}, \hat{\mathbf{f}}_{enc}^t) \leftarrow \text{Feature alignment \& Normalize}\bigl(\mathbf{f}_{enc}, \mathbf{f}_{enc}^t\bigr)$
    \STATE $\mathcal{L}_{\mathrm{Con}} \leftarrow \text{InfoNCE}\bigl(\hat{\mathbf{f}}_{enc}, \hat{\mathbf{f}}_{enc}^t\bigr)$ 
        \COMMENT{Contrastive loss for matching vs. non-matching pairs}
    \STATE \text{Update alignment layers using } $\mathcal{L}_{\mathrm{Con}}$

\STATE \textbf{Multi-modal Decoding \& Caption Generation:}
    \STATE $\hat{C} \leftarrow \text{TransformerDecoder}(\mathbf{f}_{enc})$ 
        \COMMENT{Cross-attention over $\mathbf{f}_{enc}$, autoregressive generation}

\STATE \textbf{Joint Optimization Objective:}
    \STATE $\mathcal{L}_{\mathrm{Cap}} \leftarrow \text{CrossEntropy}\bigl(\hat{C}, C_{gt}\bigr)$ 
        \COMMENT{Caption generation loss}
    \STATE $\mathcal{L}_{\mathrm{Total}} \leftarrow \mathcal{L}_{\mathrm{Cap}} + \lambda \cdot \mathcal{L}_{\mathrm{Con}}$
\end{algorithmic}
\end{algorithm}

\vspace{-0.4cm}

\begin{table*}
\begin{center}
    \caption{
    Comparison of various methods on the ScanRefer dataset \cite{chen2020scanrefer}. We evaluate the performance of each method, with and without additional 2D input, at IoU thresholds of 0.25 and 0.5. Metrics include CIDEr (C)~\cite{cider2015}, BLEU-4 (B-4)~\cite{bleu2002}, METEOR (M)~\cite{meteor2005}, and ROUGE-L (R)~\cite{rouge2004}. Our proposed 3D CoCa achieves state-of-the-art results across all settings.
    }
    \vspace{-0.3cm}
    \resizebox{\linewidth}{!}{
    \begin{tabular}{ccccccccccccccccccccc}
    \toprule
    \multirow{3}{*}{Method} &  & \multicolumn{9}{c}{w/o additional 2D input} &  & \multicolumn{9}{c}{w/ additional 2D input} \\
                            &  & \multicolumn{4}{c}{IoU = 0.25} &  & \multicolumn{4}{c}{IoU = 0.50} &  & \multicolumn{4}{c}{IoU = 0.25} &  & \multicolumn{4}{c}{IoU = 0.50} \\ \cline{3-6} \cline{8-11} \cline{13-16} \cline{18-21} 
                            &  & C$\uparrow$ & B-4$\uparrow$ & M$\uparrow$ & R$\uparrow$ &  & C$\uparrow$ & B-4$\uparrow$ & M$\uparrow$ & R$\uparrow$ &  & C$\uparrow$ & B-4$\uparrow$ & M$\uparrow$ & R$\uparrow$ &  & C$\uparrow$ & B-4$\uparrow$ & M$\uparrow$ & R$\uparrow$ \\ \hline
    Scan2Cap \cite{scan2cap_2021}                &  & 53.73       & 34.25         & 26.14       & 54.95       &  & 35.20       & 22.36         & 21.44       & 43.57       &  & 56.82       & 34.18         & 26.29       & 55.27       &  & 39.08       & 23.32         & 21.97       & 44.78       \\
    MORE \cite{MORE_2022}                        &  & 58.89       & 35.41         & 26.36       & 55.41       &  & 38.98       & 23.01         & 21.65       & 44.33       &  & 62.91       & 36.25         & 26.75       & 56.33       &  & 40.94       & 22.93         & 21.66       & 44.42       \\
    SpaCap3d \cite{spa2cap2022}                   &  & 58.06       & 35.30         & 26.16       & 55.03       &  & 42.76       & 25.38         & 22.84       & 45.66       &  & 63.30       & 36.46         & 26.71       & 55.71       &  & 44.02       & 25.26         & 22.33       & 45.36       \\
    3DJCG \cite{3djcg2022}                        &  & 60.86       & 39.67         & 27.45       & 59.02       &  & 47.68       & 31.53         & 24.28       & 51.80       &  & 64.70       & 40.17         & 27.66       & 59.23       &  & 49.48       & 31.03         & 24.22       & 50.80       \\
    D3Net \cite{chen2021d3net}                     &  & -           & -             & -           & -           &  & -           & -             & -           & -           &  & -           & -             & -           & -           &  & 46.07       & 30.29         & 24.35       & 51.67       \\
    3D-VLP \cite{3dvlp2024}                        &  & 64.09       & 39.84         & 27.65       & 58.78       &  & 50.02       & 31.87         & 24.53       & 51.17       &  & 70.73       & 41.03         & 28.14       & 59.72       &  & 54.94       & 32.31         & 24.83       & 51.51       \\
    Vote2Cap-DETR \cite{vote2cap2023}              &  & 71.45       & 39.34         & 28.25       & 59.33       &  & 61.81       & 34.46         & 26.22       & 54.40       &  & 72.79       & 39.17         & 28.06       & 59.23       &  & 59.32       & 32.42         & 25.28       & 52.53       \\
    Unit3D \cite{unit3d2023}                       &  & -           & -             & -           & -           &  & -           & -             & -           & -           &  & -           & -             & -           & -           &  & 46.69       & 27.22         & 21.91       & 45.98       \\
    Vote2Cap-DETR++ \cite{vote2cap++2024}          &  & 76.36       & 41.37         & 28.70       & 60.00       &  & 67.58       & 37.05         & 26.89       & 55.64       &  & 77.03       & 40.99         & 28.53       & 59.59       &  & 64.32       & 34.73         & 26.04       & 53.67       \\ \midrule \rowcolor{ACMLightBlue}
    3D CoCa (Ours)                             &  & \textbf{85.42}       & \textbf{45.56}         & \textbf{30.95}       & \textbf{61.98}       &  & \textbf{77.13}       & \textbf{41.23}         & \textbf{28.52}       & \textbf{57.40}       &  & \textbf{86.12}       & \textbf{44.79}         & \textbf{30.75}       & \textbf{61.45}       &  & \textbf{74.52}       & \textbf{38.42}         & \textbf{28.03}       & \textbf{55.23}       \\
    \bottomrule
    \end{tabular}
    }
    \label{exp:comparison_on_scanrefer}
\end{center}
\vspace{-0.3cm}
\end{table*}

\section{Experiments}
\subsection{Datasets and Evaluation Metrics}
\subsubsection{Datasets}
We analyze the performance of 3D captioning using two benchmark datasets: ScanRefer \cite{chen2020scanrefer} and Nr3D \cite{achlioptas2020referit_3d}. These datasets provide extensive descriptions of 3D scenes and objects generated by humans. ScanRefer contains 36,665 descriptions covering 7,875 objects in 562 scenes, while Nr3D contains 32,919 descriptions of 4,664 objects in 511 scenes. The training data for both datasets come from the ScanNet \cite{dai2017scannet} database, which contains 1,201 3D scenes. For evaluation, we use 9,508 descriptions of 2,068 objects in 141 scenes from ScanRefer and 8,584 descriptions of 1,214 objects in 130 scenes from Nr3D, all of which are from the 312 3D scenes in the ScanNet validation set.

\subsubsection{Evaluation metrics}
We use four metrics to evaluate model performance: CIDEr \cite{cider2015} measures human-like consensus via TF-IDF weighted n-gram similarity, BLEU-4 \cite{bleu2002} evaluates the accuracy of n-gram overlap between generated and reference captions, METEOR \cite{meteor2005} evaluates semantic alignment by considering synonyms and paraphrases, and ROUGE-L \cite{rouge2004} evaluates structural similarity based on the longest common subsequence, denoted as C, B-4, M, and R, respectively. Following previous studies \cite{3djcg2022,vote2cap2023,scan2cap_2021,MORE_2022,spa2cap2022}, Non-Maximum Suppression (NMS) is initially applied to filter out duplicate object predictions in the proposals. In order to accurately evaluate the model's caption generation capabilities, we adopt the $m@kIOU$ metric and set the IoU thresholds to 0.25 and 0.5 in our experiments, following~\cite{scan2cap_2021}:
\begin{equation}
    m@kIOU=\frac{1}{N}\sum_{i=1}^{N}m\left ( \hat{c}_i,C_i  \right ) \cdot \mathbb{I}\left \{ IoU\left ( \hat{b}_i,b_i  \right ) \ge k \right \},
\end{equation}
where $N$ represents the total number of annotated objects in the evaluation dataset, $\hat{c}_i$ is the generated caption, $C_i$ is the ground-truth caption, and $m$ can be any natural language generation metric, such as CIDEr \cite{cider2015}, METEOR \cite{meteor2005}, BLEU-4 \cite{bleu2002}, and ROUGE-L \cite{rouge2004}.

\subsection{Implementation Details}
We provide implementation details of different baselines. ``w/o additional 2D'' means that the input $\mathcal{P}\in \mathbb{R}^{40,000\times 10}$ contains the absolute positions of $40,000$ points representing the 3D scene, as well as \textit{color}, \textit{normal}, and \textit{height}. ``additional 2D'' means that we replace the color information with 128-dimensional \textit{multiview} features extracted from 2D images by ENet \cite{enet2020} following \cite{scan2cap_2021}. The 3D scene encoder backbone is based on EPCL \cite{huangepcl}, integrated with the frozen CLIP visual encoder \cite{clip2021}, and the text embedding is obtained by the frozen CLIP text encoder.

We train the models for 1,080 epochs using the standard cross-entropy loss and contrastive loss on ScanRefer~\cite{chen2020scanrefer} and Nr3D~\cite{achlioptas2020referit_3d}, using the AdamW optimizer~\cite{loshchilov2018decoupled} with a learning rate of $0.1$, a batch size of 4, and a cosine annealing learning rate scheduler. All experiments mentioned above are conducted on a single RTX4090 GPU.

\begin{figure*}[htbp]
    \centering
    \includegraphics[width=\linewidth]{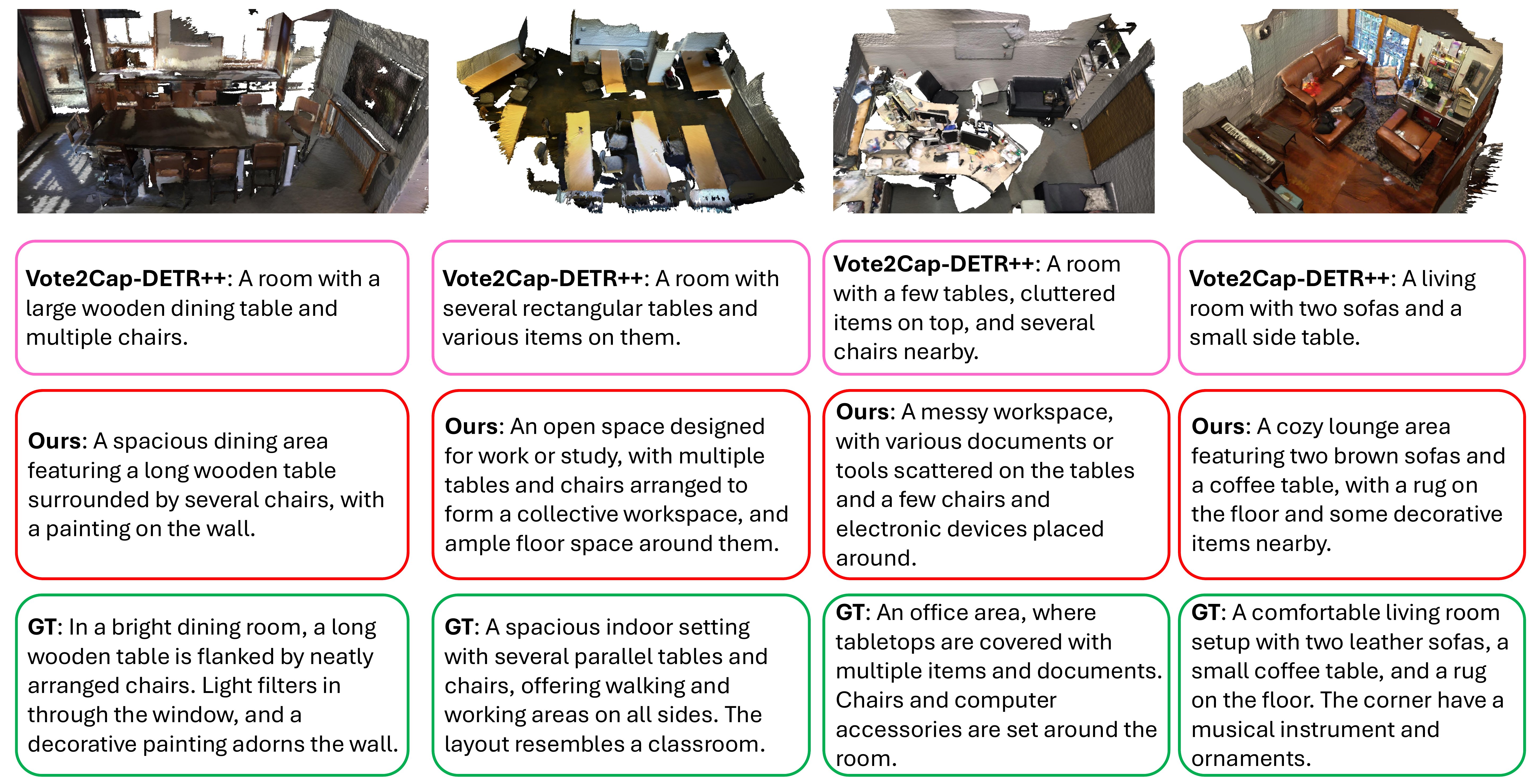}
    \vspace{-0.3cm}
    \caption{A visual comparison on the ScanRefer~\cite{chen2020scanrefer} dataset showcasing indoor scenes described by Vote2Cap-DETR++~\cite{vote2cap++2024}, our method (Ours), and the ground truth (GT), highlighting differences in descriptive accuracy and style.}
    \label{fig:visualization}
    \vspace{-0.2cm}
\end{figure*}

\subsection{Comparative Study}

\begin{table}
    \centering
    \caption{Comparison on Nr3D~\cite{achlioptas2020referit_3d} at IoU=0.5. Our model outperforms existing methods, demonstrating higher CIDEr (C)~\cite{cider2015}, BLEU-4 (B-4)~\cite{bleu2002}, METEOR (M)~\cite{meteor2005}, and ROUGE-L (R)~\cite{rouge2004} scores.}
    \vspace{-0.3cm}
    \resizebox{\linewidth}{!}{
    \begin{tabular}{ccccc}
    \toprule
    Method            & C@0.5$\uparrow$ & B-4@0.5$\uparrow$ & M@0.5$\uparrow$ & R@0.5$\uparrow$ \\ \hline
    Scan2Cap \cite{scan2cap_2021}          & 27.47           & 17.24             & 21.80           & 49.06           \\
    SpaCap3d \cite{spa2cap2022}                               & 33.71           & 19.92             & 22.61           & 50.50           \\
    D3Net \cite{chen2021d3net}                                  & 33.85           & 20.70             & 23.13           & 53.38           \\
    3DJCG \cite{3djcg2022}                                  & 38.06           & 22.82             & 23.77           & 52.99           \\
    Vote2Cap-DETR \cite{vote2cap2023}                                 & 43.84           & 26.68             & 25.41           & 54.43           \\
    Vote2Cap-DETR++ \cite{vote2cap++2024}                                  & 47.08           & 27.70             & 25.44           & 55.22           \\ \midrule \rowcolor{ACMLightBlue}
    3D CoCa (Ours)                                   & \textbf{52.84}  & \textbf{29.29}    & \textbf{25.55}  & \textbf{56.43}  \\ \bottomrule
    \end{tabular}
    }
    \label{exp:comparison_on_nr3d}
    \vspace{-0.6cm}
\end{table}

In this section, we compare the performance with existing works on metrics C, M, B-4, R as abbreviations for CIDEr~\cite{cider2015}, METEOR~\cite{meteor2005}, BLEU-4~\cite{bleu2002}, Rouge-L~\cite{rouge2004} under IoU thresholds of 0.25, 0.5 for ScanRefer (Table ~\ref{exp:comparison_on_scanrefer}) and 0.5 for Nr3D (Table~\ref{exp:comparison_on_nr3d}). In both tables, "-" indicates that neither the original paper nor any follow-up works provide such results.
\subsubsection{Scanrefer} The description in ScanRefer includes the attributes of the object and its spatial relationship with surrounding objects. As shown in Table \ref{exp:comparison_on_scanrefer}, our method outperforms the existing methods in all data settings and IoU thresholds.
\subsubsection{Nr3D} The Nr3D dataset evaluates the model’s ability to interpret human-spoken, free-form object descriptions. As shown in Table \ref{exp:comparison_on_nr3d}, our approach achieves significant performance improvements over existing models in generating diverse descriptions.

\vspace{-0.2cm}
\subsection{Ablation Study}
\vspace{-0.2cm}

\subsubsection{Contrastive learning loss impact analysis}
\begin{table}[htbp]
    \centering
    \caption{
    The impact of Contrastive Learning Loss weight $\lambda$ on the model description performance. Four evaluation indicators, CIDEr(C)~\cite{cider2015}, BLEU-4(B-4)~\cite{bleu2002}, METEOR(M)~\cite{meteor2005}, and ROUGE-L(R)~\cite{rouge2004} are listed.
    }
    \vspace{-0.3cm}
    \resizebox{\linewidth}{!}{
    \begin{tabular}{ccccc}
        \toprule
        $\lambda$ (Contrastive Weight) & C@0.5$\uparrow$ & B-4@0.5$\uparrow$ & M@0.5$\uparrow$ & R@0.5$\uparrow$ \\
        \midrule
        0   & 74.12 & 40.98 & 27.45 & 58.76 \\
        0.1 & 77.30 & 41.80 & 28.10 & 59.60 \\
        0.5 & 79.55 & 42.55 & 28.75 & 60.40 \\ \rowcolor{ACMLightBlue}
        1.0 & 85.42 & 45.56 & 30.95 & 61.98 \\
        2.0 & 76.89 & 41.50 & 28.00 & 59.30 \\
        \bottomrule
    \end{tabular}
    }
    \label{tab:ablation_contr}
    \vspace{-0.3cm}
\end{table}
We first investigate the impact of using contrastive learning loss and the sensitivity to different weight coefficients$(\lambda)$. By controlling the contrastive loss weight coefficient $\lambda = \left\{0, 0.1, 0.5, 1.0, 2.0 \right\}$, the performance of the model was compared without contrastive learning and with different strength contrastive learning strategies. As shown in Table~\ref{tab:ablation_contr}, it can be seen that when contrastive loss is not used, the model performs the worst in all indicators; the performance is significantly improved after moderate introduction of contrastive learning. For example, when $\lambda$ increases from 0 to 0.5, CIDEr increases from 74.12\% to 79.55\%, and the best performance is achieved when $\lambda$=1. However, after increasing the weight to 2.0, the indicator dropped slightly, which is still better than in the case without contrast loss. The above results show that an appropriate amount of contrast learning objectives can improve the model's ability to align and capture the semantics of 3D scenes, thereby improving the description quality.

\subsubsection{Point cloud encoder structure analysis}
\begin{table}[htbp]
\vspace{-0.3cm}
    \centering
    \caption{
    Comparison of the impact of different 3D point cloud encoder architectures on description performance. ``EPCL'' is the encoder proposed in this paper, and ``PointNet++'' is the traditional point cloud encoder.
    }
    \vspace{-0.3cm}
    \resizebox{\linewidth}{!}{
    \begin{tabular}{ccccc}
        \toprule
        Encoder Architecture & C@0.5$\uparrow$ & B-4@0.5$\uparrow$ & M@0.5$\uparrow$ & R@0.5$\uparrow$ \\
        \midrule
        PointNet++ (Baseline) & 72.48 & 38.95 & 26.80 & 56.30 \\  \rowcolor{ACMLightBlue} 
        EPCL (Proposed) & 85.42 & 45.56 & 30.95 & 61.98 \\
        \bottomrule
    \end{tabular}
    }
    \label{tab:ablation_encoder}
    \vspace{-0.3cm}
\end{table}
We compared the performance difference between the proposed EPCL point cloud encoder fused with CLIP features and the traditional PointNet++~\cite{qi2017pointnet++} point cloud encoder under the same settings. From Table~\ref{tab:ablation_encoder}, it can be seen that when using our EPCL-based encoder, the model performance is significantly better than that of PointNet++, for example, CIDEr exceeds PointNet++ by 12.94\%. The comprehensive improvement of various indicators shows that the EPCL framework combined with the pre-trained CLIP visual features effectively enhances the semantic expression and spatial modeling capabilities of point clouds and can capture richer scene information, thereby generating more accurate and detailed descriptions.
\subsubsection{Decoder architecture comparison}
\begin{table}[htbp]
    \centering
    \caption{
    The impact of different caption generation decoders on model performance. Comparison of the description indicators of the original GPT-2 generator and the CoCa-style multimodal decoder in this paper under the same visual features.
    }
    \resizebox{\linewidth}{!}{
    \begin{tabular}{ccccc}
        \toprule
        Caption Decoder & C@0.5$\uparrow$ & B-4@0.5$\uparrow$ & M@0.5$\uparrow$ & R@0.5$\uparrow$ \\
        \midrule
        GPT-2 Captioner (Baseline) & 76.20 & 41.00 & 27.80 & 59.50 \\ \midrule \rowcolor{ACMLightBlue}
        CoCa Transformer (Proposed) & 85.42 & 45.56 & 30.95 & 61.98 \\
        \bottomrule
    \end{tabular}
    }
    \label{tab:ablation_decoder}
    \vspace{-0.3cm}
\end{table}
Finally, we analyze the impact of the caption generation decoder structure on performance while keeping the output features of the visual encoder unchanged. We replace the CoCa-style multimodal Transformer decoder with the traditional GPT-2 text generation model. As shown in Table~\ref{tab:ablation_decoder}, it can be seen that the model description quality is significantly reduced when using the GPT-2 captioner. This demonstrates that the CoCa-style Transformer decoder in our approach can more effectively incorporate contrastively learned aligned visual features into the language generation process, resulting in descriptions that are more semantically rich and more closely related to the scene.

\subsection{Qualitative Results}

We compare qualitative results with the state-of-the-art Vote2Cap-DETR++ model~\cite{vote2cap++2024} in Figure~\ref{fig:visualization}. It can be seen that our method can accurately describe the attributes and categories of 3D scenes.

\section{Conclusion}
In this work, we propose 3D CoCa, a unified contrastive-captioning framework for 3D vision-language tasks. By jointly learning contrastive 3D-text representations and caption generation within a single model, 3D CoCa eliminates the need for any explicit 3D object detectors or proposal stages. This unified approach enables direct 3D-to-text alignment in a shared feature space, leading to improved spatial reasoning and more precise semantic grounding compared to previous methods. Experiments on two widely used datasets validate that our proposed 3D CoCa model significantly outperforms existing methods across standard captioning metrics and proves the benefits of our contrastive learning strategy.


\end{document}